%% file: main.tex
\documentclass{article}
\usepackage{./style/ijcai17}

\usepackage{times}
\usepackage{theorem}
\usepackage[dvipsnames]{xcolor} 
\usepackage{xspace}
\usepackage{epsfig}
\usepackage{graphicx}
\usepackage{amsmath}
\usepackage{amssymb}
\usepackage{comment}
\usepackage [autostyle, english = american]{csquotes}
\MakeOuterQuote{"}
\def\st{\emph{s.t}\onedot} 
\usepackage{array}
\usepackage{adjustbox}
\usepackage{url}
 
\makeatletter
\DeclareRobustCommand\onedot{\futurelet\@let@token\@onedot}
\def\@onedot{\ifx\@let@token.\else.\null\fi\xspace}

\def\eg{\emph{e.g}\onedot} 
\def\ie{\emph{i.e}\onedot} 
\def\cf{\emph{c.f}\onedot}

\def\etal{\emph{et al}\onedot}
\makeatother
\usepackage[font={small}]{caption}

\begin{document}

\input{./sections/macros}

%%%%%%%%% TITLE

\title{How a General-Purpose Commonsense Ontology can Improve Performance of Learning-Based Image Retrieval\thanks{
This work was partially funded by grants FONDECYT 1151018 and 1150328. 
Rodrigo also gratefully acknowledges funding from CONICYT (Becas Chile).
We thank the IJCAI-17 reviewers and Margarita Castro for providing valuable insights into this research.
%The first author also gratefully acknowledges funding from CONICYT (Becas Chile) 
%and thanks Margarita Castro for providing valuable insights into this research.
}
}
% This removes the 'thanks mark' from the title
%\renewcommand\footnotemark{}

\author{
Rodrigo Toro Icarte\textsuperscript{$\dagger$}, Jorge A. Baier\textsuperscript{$\ddagger$,$\S$}, Cristian Ruz\textsuperscript{$\ddagger$}, Alvaro Soto\textsuperscript{$\ddagger$} \\
\textsuperscript{$\dagger$}University of Toronto, \textsuperscript{$\ddagger$}Pontificia Universidad Cat\'olica de Chile, \textsuperscript{$\S$}Chilean Center for Semantic Web Research \\
rntoro@cs.toronto.edu, \{jabaier, cruz, asoto\}@ing.puc.cl
}

\maketitle

\begin{abstract}
The knowledge representation community has built general-purpose ontologies 
which contain large amounts of commonsense knowledge over relevant aspects of 
the world, including useful visual information, e.g.: ``a 
ball is \emph{used by} a football player'', ``a tennis player is \emph{located 
at} a tennis court''. Current state-of-the-art approaches for visual 
recognition do not exploit these rule-based knowledge sources. Instead, they 
learn recognition models directly from 
training examples. In this paper, we study how general-purpose 
ontologies---specifically, MIT's ConceptNet ontology---can improve the 
performance of state-of-the-art vision systems. As a testbed, we tackle the 
problem of sentence-based image retrieval. Our retrieval approach incorporates 
knowledge from ConceptNet on top of %the output of 
a large pool of object 
detectors derived from a deep learning technique. In our experiments, we show 
that ConceptNet can improve performance on a common benchmark dataset. Key to 
our performance is the use of the \esp dataset to select visually relevant
relations from ConceptNet. Consequently, a main conclusion 
of this work is that 
general-purpose commonsense ontologies improve performance
on visual reasoning tasks when properly filtered to select meaningful visual 
relations.
\end{abstract}

%\begin{abstract}
%  The knowledge representation community has invested great efforts in
%  building general-purpose ontologies which contain large amounts of
%  commonsense knowledge on various aspects of the world. Among the
%  thousands of assertions contained in them, many express relations
%  that can be regarded as visual; e.g., ``a ball is \emph{used by} a
%  football player'', ``a tennis player is \emph{located at} a tennis
%  court''. Being learning-based, current state-of-the-art approaches
%  are not capable of exploiting existing knowledge written in the form
%  of rules but rather extract all visual information directly from
%  (plenty of) learning examples. In this paper we study the question
%  of whether or not general-purpose ontologies---specifically, MIT's
%  ConceptNet ontology---could play a role in state-of-the-art vision
%  systems, as it seems plausible in principle that general-knowledge
%  may be complementary to knowledge acquired from examples. We
%  identify a number of issues that make this integration non-trivial
%  and challenging, among which is the existence of many ``noisy
%  relations'' in ConceptNet. We propose approaches that deal with
%  these hurdles to some extent, identifying aspects that could be
%  improved by future work. We describe an integration of ConceptNet
%  with Microsoft XXX YYY detectors, and propose an approach to image
%  retrieval that improves upon the original detector-based approach,
%  leading to state-of-the-art results (maybe be specific about numbers
%  here). Say sth more about results?
%\end{abstract}

% Squeezing

\setlength\textfloatsep{0.5cm} 
\setlength\abovecaptionskip{0.15cm}
\setlength\dbltextfloatsep{0.1cm}
\setlength\theorempreskipamount{2pt plus 1pt minus 1pt}
\setlength\theorempostskipamount{1pt plus 1pt minus 1pt}
\renewcommand{\topfraction}{0.95}
\renewcommand{\textfraction}{0.05}
\renewcommand{\floatpagefraction}{0.95}

%%%%%%%%% BODY TEXT
\input{./sections/introduction.tex}

\input{./sections/previous_work.tex}

\input{./sections/preliminaries.tex}

\input{./sections/baseline.tex}

\input{./sections/methodology.tex}
\input{./sections/results.tex}

%\input{results_appendix.tex}
%\input{word_figure.tex}

\input{./sections/conclusions.tex}

%\input{./sections/acknowledgments.tex}

\bibliographystyle{./style/named} 

{\normalsize
\bibliography{./refs/finalShort}
}
%\bibliography{./refs/AsaReferences,./refs/ref,./refs/results}

\end{document}

%% file: sections/macros.tex
\newcommand{\word}[1]{\emph{#1}}
\newcommand{\angled}[1]{\ensuremath{\langle #1\rangle}}
\newcommand{\oneNscore}{\ensuremath{1\text{-} N\text{-}Score}\xspace}
\newcommand{\C}[1]{\ensuremath{\mathcal{#1}}\xspace}
\newcommand{\Paragraph}[1]{\smallskip\noindent\textbf{#1}~}
\newcommand{\CN}{ConceptNet\xspace}
\newcommand{\rel}[1]{--\textit{#1}\ensuremath{\rightarrow}}
\newcommand{\fs}[1]{\fontsize{#1}{#1}\selectfont}
\newcommand{\MIL}{\ensuremath{\mathrm{MIL}}\xspace}
\newcommand{\MILSTEM}{\ensuremath{\mathrm{MILSTEM}}\xspace}
\newcommand{\CNs}{\ensuremath{\mathrm{CN}}\xspace}
\newcommand{\Phat}{\ensuremath{\hat{P}}\xspace}
\newcommand{\phat}{\ensuremath{\hat{p}}\xspace}
\newcommand{\cn}{\ensuremath{\mathit{cn}}\xspace}
\newcommand{\stDet}{\ensuremath{\mathit{stDet}}\xspace}
\newcommand{\cnDet}{\ensuremath{\mathit{cnDet}}\xspace}
\newcommand{\mean}{\operatorname{mean}}
\newcommand{\meanA}{\operatorname{mean-a}}
\newcommand{\meanG}{\operatorname{mean-g}}
\newcommand{\eqdef}{\ensuremath{\overset{\text{def}}{=}}}
\newcommand{\esp}{\textsc{espgame}\xspace}

\newcommand{\citea}[1]{\citeauthor{#1}~\shortcite{#1}}

% Colors
\definecolor{mil}{HTML}{008000}
\definecolor{stem}{HTML}{AAA839}
\definecolor{cn}{HTML}{A00000}

%% file: sections/introduction.tex
\section{Introduction}
The knowledge representation community has recognized that commonsense 
knowledge bases are needed for reasoning in the real world. Cyc 
\cite{Lenat:1995} and ConceptNet  (CN) \cite{Havasi:EtAl:2007} are two 
well-known examples of large, publicly available commonsense knowledge bases.
% In particular, in this work we focus on ConceptNet (CN), a large hyper-graph 
%containing information about millions of world concepts and their relations, 
%such as, ``books can be made from paper''.

CN has been used successfully for tasks that require rather complex commonsense 
reasoning, including a recent study that showed that the information in CN may 
be used to score as good as a four-year old in an IQ test \cite{OhlssonSTU13}. 
CN also contains many assertions that seem visually relevant; such as \emph{``a 
chef is (usually) located at the kitchen''}.

State-of-the-art approaches to visual recognition tasks are mostly based on 
learning techniques. Some use mid-level representations 
\cite{Singh:EtAl:2012,LobelA:EtAl:2013}, others 
deep hierarchical layers of composable features 
\cite{LeCun:EtAl:2007,Krizhevsky:EtAl:12}. Their goal is to uncover 
visual spaces where visual similarities carry enough information to achieve 
robust visual recognition. While some approaches exploit knowledge and semantic
information 
\cite{Liu11Attributes,Espinace:EtAl:2013}, none of 
them utilize large-scale 
ontologies to improve performance.

In terms of CN, previous works have suggested 
that incorporating CN knowledge to visual applications is nontrivial 
\cite{LeUB13,XieH13,Snoek2007}. Indeed, poor results in \cite{LeUB13} and 
\cite{XieH13} can be attributed to a non-negligible rate of noisy relations 
in CN. The work in \cite{Snoek2007} helps to support this claim: ``...manual
process (of CN relations) guarantees high quality links, which are necessary
to avoid obscuring the experimental results.''

\begin{figure}[t]
\centering 
\includegraphics[width=0.98\linewidth]{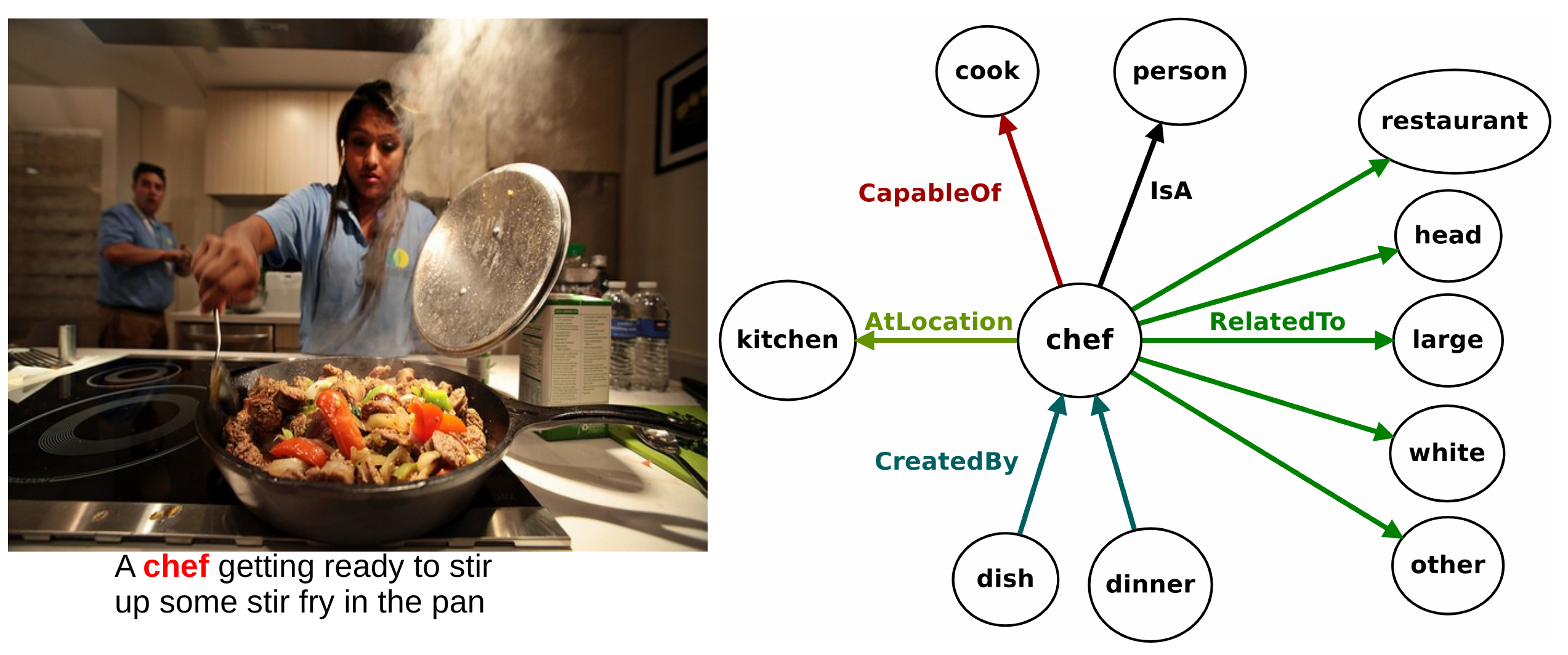}
%\fbox{\rule{0pt}{2in} \rule{0.9\linewidth}{0pt}}
%\includegraphics[height=0.28\linewidth]{jc/fig1_spatial.pdf}
%&
%\includegraphics[height=0.28\linewidth]{jc/fig1_temporal.pdf}\\
%(a) Spatial & (b) Temporal composition
\vspace{-0.2cm}
\caption{\textbf{Left.} An image and one of its associated sentences from the MS 
COCO dataset. Among its words, the sentence features the word \textit{Chef}, 
for which there is not a visual detector available. \textbf{Right.} Part of the 
hypergraph at distance 1 related to word \textit{Chef} in ConceptNet. In the 
list of nodes related to the concept \textit{Chef}, there are several 
informative concepts for which we have visual detectors available.}
\label{fig:frontfigure}
\end{figure}

In this paper we study the question of how large and publicly 
available general-purpose commonsense knowledge repositories, specifically CN, 
can be used to improve state-of-the-art vision techniques. We focus on the 
problem of sentence-based image retrieval. We approach the problem by assuming 
that we have visual detectors for a number of words, and describe a CN-based method 
to enrich the existing set of detectors. Figure \ref{fig:frontfigure} shows an 
illustrative example: An image retrieval query contains the word 
\textit{Chef}, for which there is not a visual detector available. In this 
case, the information contained in the nodes directly connected to the concept 
\textit{Chef} in CN provides key information to trigger related visual 
detectors, such as \textit{Person}, \textit{Dish}, and \textit{Kitchen} 
%that results key to retrieve the intended image.
that are highly relevant to retrieve the intended image.

Given a word $w$ for which we do not have a visual detector available, we 
propose various probabilistic-based approaches to use CN's relations to estimate 
the likelihood that there is an object for $w$ in a given image. Key to the 
performance of our approach is an additional step that uses a complementary 
source of knowledge, the \esp dataset \cite{Ahn:Dabbish:2005}, to filter out noisy 
and non-visual relations provided by CN. Consequently, a main 
conclusion of this work is that filtering out relations from 
CN is very important for achieving good performance, suggesting that future work 
that attempts to integrate pre-existing general knowledge with machine learning 
techniques should put close attention to this issue.

The rest of the paper is organized as follows: Section
\ref{Sec:PreviousWork} reviews related work; Section
\ref{Sec:Preliminaries} describes the elements used in this paper;
Sections \ref{Sec:Baseline} and \ref{Sec:Method} motivate and describe
our proposed method; Section \ref{Sec:Results} presents qualitative
and quantitative experiments on standard benchmark datasets; finally,
Section \ref{Sec:Conclusions} presents future
research directions and concluding remarks.

%% file: sections/previous_work.tex
\section{Previous Work}\label{Sec:PreviousWork} 
The relevance of contextual or semantic information to visual recognition has
long been acknowledged and studied by the cognitive psychology and
computer vision communities \cite{Biederman:1972}. In
computer vision, the main focus has been on using contextual relations in the 
form of object co-occurrences,
and geometrical and spatial constraints. Due to space constraints, we refer the reader to
\cite{Marques:EtAl:2011} for an in-depth
review about these topics. As a common issue, these methods do not employ
high-level semantic relations as the one included in CN. 

Knowledge acquisition is one of the main challenges of using a semantic based 
approach to object recognition. One common approach to obtain this knowledge is 
via text mining \cite{Rabinovich:EtAl:2007,Espinace:EtAl:2013} or crowd 
sourcing 
\cite{imagenet_cvpr09}. As an alternative, recently, 
\citea{NEIL:2013} and \citea{LEVAN:2014} present bootstrapped approaches where an 
initial set of object detectors and relations is used to mine the web in order 
to discover new object instances and new commonsense relationships. The new 
knowledge is in turn used to improve the search for new classifiers and semantic 
knowledge in a never ending process. While this strategy opens new
opportunities, unfortunately, as it has been pointed out by
\citea{Ahn:Dabbish:2005}, public information is biased. In particular, 
commonsense knowledge is so obvious that it is commonly tacit and not
explicitly included in most information sources. Furthermore, unsupervised or
semi-supervised semantic knowledge extraction techniques often suffer from
semantic drift problems, where slightly misleading local association are
propagated to lead to wrong semantic inference. 

%Recently, work on automatic image captioning has made great advances to 
%integrate image and text data 
%\cite{karphati:Caption:2015,Bengio:Caption:2015}. 
%These approaches use datasets consisting of images as 
%well as sentences describing their content, such as the Microsoft COCO dataset 
%\cite{LinMBHPRDZ14}. Coincidentally, all these methods share similar ideas which 
%follow initial work in \cite{Bengio:Wasabi:2010}. Briefly, these works employ 
%deep neural network models, mainly convolutional and recurrent neural networks, 
%to infer a suitable alignment between sentence snippets and the corresponding 
%image region that they describe. In contrast to our proposed approach, these 
%methods do not make explicit use of high level semantic knowledge.

Recently, work on automatic image captioning has made great advances to 
integrate image and text data 
\cite{karphati:Caption:2015,Bengio:Caption:2015,KleinLSW15}. 
These approaches use datasets consisting of images as 
well as sentences describing their content, such as the Microsoft COCO dataset 
\cite{LinMBHPRDZ14}. Coincidentally, work by \citea{karphati:Caption:2015,Bengio:Caption:2015} 
share similar ideas which follow initial work by \citea{Bengio:Wasabi:2010}. 
Briefly, these works employ deep neural network models, mainly convolutional and 
recurrent neural networks, to infer a suitable alignment between sentence snippets 
and the corresponding image region that they describe. \cite{KleinLSW15}, 
on the other hand, proposes to use the Fisher Vector as a sentence representation 
instead of recurrent neural networks.
In contrast to our approach, these 
methods do not make explicit use of high level semantic knowledge.

In terms of works that use ontologies to perform visual 
recognition, \cite{MaillotT08} builds a custom ontology to perform visual 
object recognition. \cite{Ordonez:EtAl:2015} uses Wordnet and a large set of 
visual object detectors to automatically predict natural nouns that people will 
use to name visual object categories. \cite{FeiFeiAffordance:ECCV:14} 
uses Markov Logic Networks and a custom ontology %harvested from different sources, 
to identify several properties related to object 
affordance in images. In contrast to our work, these methods target different 
applications. Furthermore, they do not exploit the type of commonsense 
relations that we want to extract from CN.

%In this paper, we pioneer the
%adaptation of public commonsense
%knowledge bases, such as ConceptNet, to support unconstrained visual recognition
%tasks. As shown in \cite{AAAIW137023}, according to a standard test, ConceptNet
%matches the average Verbal IQ of a 4-year-old child. In particular, one of the
%main results of this paper is to show that knowledge extracted from an ontology
%of common sense knowledge can provide superior performance that the n-grams
%relations provided by a massive textual dataset like the one provided by google
%API.

%% file: sections/preliminaries.tex
\section{Preliminaries}
\label{Sec:Preliminaries}
%Below we present an overview of the elements we use in this paper,
%including the ontology we use and background on the image
%retrieval task.

\Paragraph{ConceptNet}
\CN (CN) \cite{Havasi:EtAl:2007} is a commonsense-knowledge semantic network
which represents knowledge in a hypergraph structure. Nodes in the
hypergraph correspond to a concept represented by a word or a
phrase. In addition, hyperarcs  represent relations between
nodes, and are associated with a weight that expresses the confidence
in such a relation. As stated in its webpage, CN is a knowledge base ``containing
lots of things computers should know about the world, especially when
understanding text written by people.''

%CN relations are commonsensical in the sense that they represent
%knowledge that is standard for most humans (see Figure~\ref{table:sofa}
%for a sample).
%One of the design principles behind CN is that
%knowledge is automatically generated from sentences coming from the
%\emph{Open Mind Comonsense Corpus} \cite{singh2002open,conceptnet},
%which are natural-language sentences created by humans. Online games
%(\eg, \emph{Verbosity} \cite{AhnKB06}) have also been used to
%% massively
%collect relations for CN.

\begin{figure}
{\fs{8}
\begin{tabular}{l|l}
  \CN relation & \CN's description \\\hline
  sofa \rel{IsA} piece of furniture &  \emph{A sofa is a piece of furniture} \\
  sofa \rel{AtLocation} livingroom & \emph{Somewhere sofas can be is livingroom}\\
  sofa \rel{UsedFor} read book & \emph{A sofa is for reading a book}\\
%  sofa \rel{RelatedTo} book & \emph{When you think about a sofa, books also comes to mind}\\
  sofa \rel{MadeOf} leather & \emph{sofas are made from leather}
\end{tabular}
}
\caption{A sample of CN relations that involve the concept
  \emph{sofa}, together with the English description provided by the
  CN team in their website.}\label{table:sofa}
\end{figure}

Among the set of relation types in CN, a number of them can be
regarded as ``visual,'' in the sense that they correspond to
relations that are important in the visual world (see Figure~\ref{table:sofa}). These include relations for spatial co-occurrence (e.g., \word{LocatedNear}, \word{AtLocation}), visual properties of objects (e.g., \word{PartOf}, \word{SimilarSize}, \word{HasProperty}, \word{MadeOf}), and actions
(e.g., \word{UsedFor}, \word{CapableOf}, \word{HasSubevent}).

%CN 5.3, the version we use in our experiments, contains over 4
%million nodes and over 13 million relations. Many of these relations
%are accurate. Indeed Singh~\emph{et al.} \cite{singh2002open}
%reported that human evaluators have rated 75\% of items as ``largely
%true'', 82\% as ``largely objective'', and 85\% as ``largely making
%sense''. 

%CN has been used successfully for tasks that require rather complex
%commonsense reasoning. A recent study showed that
%the information in CN may be used to score as good as a four-year-old
%person in an IQ test \cite{OhlssonSTU13}.

Even though CN's assertions are reasonably accurate \cite{singh2002open} and that it can be used to score as good as a four-year-old in an IQ test \cite{OhlssonSTU13}, it contains a number of so-called \emph{noisy
  relations}, which are relations that do not correspond to a true
statement about the world. Two examples of these for the concept
\emph{pen} are ``pen \rel{AtLocation} pen'', ``pig \rel{AtLocation}
pen''. The existence of these relations is an obvious hurdle when
utilizing this ontology.
%We discuss later some of the implications of
%this for visual reasoning. 

\Paragraph{Stemming} 
A standard Natural Language Processing technique used below
is \emph{stemming}. The stemming of word $w$ is an English word resulting from stripping a suffix out of $w$. It is a heuristic process that aims at returning the ``root'' of a word. For example, the stemming of the words \emph{run}, \emph{runs}, and \emph{running} all return the word \emph{run}. For a word $w$, we denote its stemmed version as $st(w)$. If $W$ is a
set of words, then $st(W)=\{st(w) : w\in W\}$. 

%\Paragraph{Sentence-Based Image Retrieval}
%Given a set of images $\mathcal{I}$ and a natural-language query
%string $t$, the objective of image retrieval is to rank the images in
%$\mathcal{I}$ according to their ``relevance'' with respect to $t$. As
%such, the problem reduces to building a function that returns a
%numeric score to each image in $I\in \mathcal{I}$ given $t$ that correlates
%with the relevance of $I$ with respect to $t$.
%Two recent state-of-the-art approaches to image retrieval are those by
%\citea{KleinLSW15}, based on Mixture Models, and that by 
%\citea{karphati:Caption:2015}, based on a bidirectional recurrent
%neural network. 

%% file: sections/baseline.tex
	% MIL PROB
\section{A Baseline for Image Retrieval }
\label{Sec:Baseline}
To evaluate our technique for image retrieval, we chose as a baseline a
simple approach based on a large set of visual word detectors 
\cite{fangCVPR15}. These detectors, that we refer to as Fang \etal's
detectors, 
were trained over the MS~COCO image
dataset \cite{LinMBHPRDZ14}. Each image in this dataset is associated with 5
natural-language descriptions. Fang \etal's detectors were trained to detect 
instances of words appearing in the sentences associated 
to MS~COCO images. As a result, they obtain a set of visual word 
detectors
for a vocabulary $V$, which contains the 1000 most common words used to
describe images on the training dataset.

Given an image $I$ and a word $w$, Fang \etal's detector outputs a
score between 0 and 1. With respect to training data, such a score can be 
seen as an estimate of the probability that image $I$ has been described
with word $w$. Henceforth,  we denote such a score
by $\Phat_V(w\mid I)$.

A straightforward, but effective way of applying these detectors to
image retrieval is by simply multiplying the scores. Specifically, given a
text query $t$ and an image $I$ we run the detectors on $I$ for words
in $t$ that are also in $V$ and multiply their output scores. We
denote this score by $MIL$ (after Multiple Instance Learning, the
technique used in \cite{fangCVPR15} to train the detectors). Mathematically,
\begin{equation} 
\MIL(t,I) = \prod_{\text{$w\in V \cap S_t$}} \Phat_V(w| I),
\label{eq:mil}
\end{equation}
where $S_t$ is the set of words in the text query $t$.

The main assumption behind Equation \ref{eq:mil} corresponds to an independence 
assumption among word detectors given an image $I$. This is similar to the Naive 
Bayes assumption used by the classifier with the same name \cite{Mitchell:1997}. 
In Section~\ref{Sec:Results}, we show that, although simple, this 
score outperforms previous works %state-of-the-art 
(\eg, \cite{KleinLSW15,karphati:Caption:2015}). 
%While these detectors were trained 
%using MS~COCO dataset, we still found---suprisingly---that they lead to 
%state-of-the-art performance when they are used in a sentence-based image 
%retrieval task using this dataset. 

%% file: sections/methodology.tex
\section{CN-based Detector Enhancement}
\label{Sec:Method}
The $\MIL$ score has two limitations. First, it considers the text
query as a set of independent words, ignoring their semantic relations and 
roles in the sentence. Second, it is
limited to the set $V$ of words the detector has been trained
for. While the former limitation may also be present in other
state-of-the-art approaches to image retrieval, the latter is inherent
to any approach that employs a set of visual word detectors for image
retrieval. 

Henceforth, given a set of words $V$ for which we have a detector, we
say that word $w$ is \emph{undetectable} with respect to $V$ iff $w$
is not in $V$, and we say $w$ is \emph{detectable} otherwise.

%In the rest of the section we describe a  technique to enhance a
%detector-based approach for image retrieval with undetectable
%words. 

%The first exploits the knowledge in CN, while the second is a
%simple extension of \MIL that uses stemming. This second approach does
%not incorporate any commonsense knowledge but we include this second
%approach for fairness, because stemming is also used in our CN-based
%approach and also wanted to explore the benefits of using stemming
%within the \MIL score.

\subsection{CN for Undetectable Words}
Our goal is to provide a score to each image analogous to that defined
in Equation \ref{eq:mil}, but including undetectable words. A
first step is to define a score for an individual undetectable word
$w$.  Intuitively, if $w$ is an undetectable word, we want an estimate
analogous to $\Phat_V(w| I)$. Formally, the problem we address can be
stated as follows: given an image $I$ and a word $w$ which is
undetectable wrt.\ $V$, compute the estimate $\Phat(w| I)$ of the
probability $P(w| I)$ of $w$ appearing in $I$.

To achieve this, we are inspired by the following: for most words
representing a concept $c$, CN ``knows'' a number of concepts
related to $c$ that share related visual
characteristics. For example, if $w$ is \emph{tuxedo}, then \emph{jacket} may provide useful information since ``tuxedo\rel{IsA} jacket'' is in
CN.
% Likewise, other visual CN relations such as \emph{AtLocation} or \emph{MadeOf} may also provide useful information. 

We define $\cn(w)$ as the set of concepts that are directly related to the stemmed
version of $w$, $st(w)$, in CN. We propose to compute $\Phat(w| I)$ based on the
estimation $\Phat(w'| I)$ of words $w'$ that appear in $\cn(w)$.
Specifically, by using standard probability theory we can write the
following identity about the \emph{actual} probability function
$P$ and every $w'\in \cn(w)$:
\begin{equation}
P(w | I) = P(w | w', I)P(w'| I) + P(w | \neg w', I)P(\neg w'| I),
\label{eq:totalprob}
\end{equation}
where $P(w| w',I)$ is the probability that there is an object in $I$
associated to $w$ given that there is an object associated to $w'$ in
$I$. Likewise $P(w|\neg w',I)$ represents the probability that there
is an object for word $w$ in $I$, given that no object associated to
word $w'$ appears in $I$. 

Equation \ref{eq:totalprob} can be re-stated in terms of
estimations. %, thus for \emph{any} $w'\in \cn(w)$,
$P(w| I)$ can be
estimated by $\phat(w',w,I)$, which is defined by 
\begin{multline}\label{eq:phatww}
\phat(w',w,I) \eqdef \\
\Phat(w | w', I)\Phat(w'| I) + \Phat(w | \neg
w', I)\Phat(\neg w'| I).
\end{multline}
However the drawback with such an approach is that it does not tell us
which $w'$ to use. Below, we propose to aggregate \phat over the set of
all concepts $w'$ that are related to $w$ in CN. Before stating such an
aggregation formally, we focus on how to compute $\Phat(w |  w', I)$
and $\Phat(w'|  I)$.

Let us define the set $\stDet(w,V)$ as the set of
words in the set $V$ such that when stemmed are equal to $st(w)$; \ie,
$\stDet(w,V)= \{w'\in V : st(w)=st(w')\}$. Intuitively, $\stDet(w,V)$
contains all words in $V$ whose detectors can be used to detect
$w$ after stemming. Now we define $\Phat(w'| I)$ as:
\begin{equation}\label{eq:Pst}
\Phat(w'| I)=\max_{w\in \stDet(w',V)} \Phat_V(w| I), 
\end{equation}
\ie, to estimate how likely it is that $w'$ is in $I$, we look for
a word $w$ in the set of $V$ whose stemmed version matches
the stemmed version of $w'$, and that maximizes $\Phat_V(w| I)$. 

Now we need to define how to compute $\Phat(w |  w', I)$. We tried two
options here. %(we report both in Section~\ref{Sec:Results}). 
The first
is to assume $\Phat(w |  w',I)=1$, for every $w,w'$. This is because for some
relation types it is correct to assume that $\Phat(w |  w',I)$ is equal
to 1. For example $\Phat(\text{person} |  \text{man},I)$ is 1 because
there is a CN relation ``man\rel{IsA} person''. While it is clear that
we should use 1 for the \emph{IsA} relation, it is not clear whether
or not this estimate is correct for other relation types. Furthermore,
since CN contains noisy relations, using 1 might yield significant
errors.

Our second option, which yielded better results, is to approximate
$\Phat(w |  w', I)$ by %equal to an estimate of 
$P(w| w')$; i.e., the probability that an image that contains an object for $w'$ contains an
object for word $w$.  $P(w| w')$ can be estimated from the \esp
database \cite{Ahn:Dabbish:2005}, which contains tags for many images. After stemming each
word, we simply count the number of images tagged in which both $w$
and $w'$ occur and divide it by the number of images tagged by $w'$.

Now we are ready to propose a CN-based estimate for $P(w| I)$ when $w$ is
undetectable. As discussed above, $P(w| I)$ could be estimated by the
expression of Equation~\ref{eq:phatww} for any concept $w'\in cn(w)$. As it
is unclear which $w'$ to choose, we propose to aggregate over
$w'\in cn(w)$ using three aggregation functions. Consequently, we identify
three estimates of $P(w| I)$ that are defined by:
\begin{equation}\label{eq:phatf}
\Phat_\C{F}(w| I)= \C{F}_{w'\in\cnDet(w,V)} \phat(w',w,I),
\end{equation}
where $\cnDet(w,V)=\{w'\in \cn(w):\stDet(w',V)\neq \emptyset\}$ is the
set of concepts related to $w$ in CN for which there is a stemming
detector in $V$, and $\C{F} \in\{\min,\max,\mean\}$.

\subsection{The CN Score}
With a definition in hand for how to estimate the score of an
individual undetectable word $w$, we are ready to define a CN-based
score for a complete natural-language query $t$. For any $w$ in $t$,
what we intuitively want is to use the set of detectors whenever $w$ is
detectable and $\Phat_\C{F}(w| I)$ otherwise.

To define our score formally, a first step is to extend the \MIL score
with stemming. Intuitively, we want to resort to detectors in $V$ as
much as possible, therefore, we will attempt to stem a word and use a
detector before falling back to our CN-based score. Formally,
\begin{equation}
\MILSTEM(t,I) = \MIL(t,I)\times \prod_{w\in W_t'} \Phat(w|  I),
\label{eq:milstem}
\end{equation}
where $W_t'$ is the set of words in $t$ that are undetectable wrt.\
$V$ but that are such that they have a detector via stemming (\ie,
such that $\stDet(w,V)\neq \emptyset$), and where $\Phat(w|  I)$ is defined
by Equation~\ref{eq:Pst}.

Now we define our CN score which depends on the aggregation function
$\C{F}$. Intuitively, we want to use our CN score with those words
that remain to be detected after using the detectors directly and
using stemming to find more detectors. Formally, let $W_t''$ be the
set of words in the query text $t$ such that (1) they are undetectable
with respect to $V$, (2) they have no stemming-based detector
(\ie. $stDet(w,V)=\emptyset$), but (3) they have at least one related
concept in CN for which there is a detector (\ie,
$cnDet(w,V)\neq \emptyset$). Then we define:
\begin{equation}
\CNs_\C{F}(t,I)=\MILSTEM(t,I)\times \prod_{w\in W_t''} \Phat_\C{F}(w|  I),
\end{equation}
for $\C{F} \in\{\min,\max,\mean\}$.

% CODE!
\Paragraph{Code} Our source code is publicly available in \url{https://bitbucket.org/RToroIcarte/cn-detectors}.

%% file: sections/results.tex
\section{Results and Discussion}
\label{Sec:Results}
We evaluate our algorithm over the MS~COCO image database 
\cite{LinMBHPRDZ14}. Each image in this set contains 5 natural-language
descriptions.  Following \cite{karphati:Caption:2015} and \cite{KleinLSW15} we
use a specific subset of 5K images and evaluate the methods on the
union of the sentences for each image. %Henceforth, 
We refer to this subset as COCO~5K.

%In our tables 
We report the mean and median rank of the \emph{ground
 truth} image; that is, the one that is tagged by the query text
being used in the retrieval task. We report also the $k$-recall
($r@k$), for $k\in\{1,5,10\}$, which corresponds to the percentage of
times the correct image is found among the top $k$ results.

Recall that we say that a word $w$ is detectable when there is a
detector for $w$. In this section we use Fang \etal's detectors, %\cite{fangCVPR15} 
which is comprised by 616 detectors for nouns, 176
%verb detectors and 119 detectors for adjectives.  In addition,
for verbs, and 119 for adjectives.  In addition,
we say that a word $w$ is stemming-detectable if it is
among the words considered by the \MILSTEM score, and we say a word is
CN-detectable if it is among the words included in the CN-score.

\subsection{Comparing Variants of CN}
The objective of our first experiment is to compare the performance
of the various versions of our approach that use different aggregation
functions. Since our approach uses data from \esp we also compare to
an analogous approach that uses only \esp data, without knowledge from
CN. This is obtained by interpreting that word $w$ is related to a
word $w'$ if both occur on the same \esp tag, and using the same
expressions presented in Section~\ref{Sec:Method}. We consider that a
comparison to this method is important because we want to evaluate
the impact of using an ontology with general-purpose knowledge versus
using a crowd-sourced, mainly visual knowledge such as that in \esp.

% Ontology focus
Table~\ref{tab:mscoco5k_subset} shows results over the maximal
subset of COCO~5K such that a query sentence has a CN-detectable word
that is not stemming-detectable, including $9,903$
queries. The table shows results for our baselines, CN\_OP and ESP\_OP
(with OP = MIN (minimum), MEAN\_G (geometric mean), MEAN\_A
(arithmetic mean) and MAX (maximum)). Results show that
algorithms based on CN perform better in all the reported metrics,
including the median rank. Overall, the MAX version of CN seems to
obtain the best results and thus we focus our analysis on it.
%we select it as the method that we compare to other approaches below.

% Hard examples plus alternatives approaches

	\begin{table}\centering
          {\fs{9}
            \renewcommand{\arraystretch}{1.2} 
	\begin{tabular}{ | l | c c c c c |}
	\hline
	\textbf{Database} & r@1 & r@5 & r@10 & median & mean  \\
	$\subset$ COCO~5K &     &     &      & rank   & rank \\
	\hline
	\textbf{Our baselines}   & & & & & \\
	\MIL & 13.2 & 33.4 & 45.2 & 13 & 82.2 \\
	\MILSTEM & 13.5 & 33.8 & 45.7 & 13 & 74.6 \\
	\hline
	%CN W1 & 13.2 & 33.7 & 46.0 & 13.0 & 66.3 \\
	%CN W & 13.4 & 34.1 & 46.7 & 13.0 & 65.1 \\
	%CN W2V & 13.3 & 33.7 & 46.1 & 13.0 & 65.6 \\
	%CN ESPGAME & 14.0 & 35.2 & 47.4 & \textbf{12.0} & 60.8 \\
	%\hline
	\textbf{Without CN}   & & & & & \\
	ESP\_MIN     & 12.6 & 30.7 & 41.1 & 17 & 122.4 \\
	ESP\_MEAN\_G & 13.5 & 34.0 & 46.0 & 13 & 70.5 \\
	ESP\_MEAN\_A & 13.6 & 34.2 & 46.2 & 13 & 69.0 \\
	ESP\_MAX     & 13.5 & 33.7 & 45.7 & 13 & 66.2 \\
	\hline
	\textbf{Using CN} & & & & & \\
	CN\_MIN      & 14.3 & 34.6 & 46.6 & \textbf{12} & 68.3 \\
	CN\_MEAN\_G  & 14.5 & 35.2 & 47.3 & \textbf{12} & 64.3 \\
	CN\_MEAN\_A  & \textbf{14.6} & 35.6 & 48.0 & \textbf{12} & 61.2 \\
	CN\_MAX      & 14.3 & \textbf{35.9} & \textbf{48.2} & \textbf{12} & \textbf{60.6} \\
	\hline	
	\end{tabular}}
	\caption{Subset of COCO~5K with sentences that contain at least one undetectable word.}
	\label{tab:mscoco5k_subset}
	\end{table}

% Qualitative examples
Figure~\ref{fig:qualitative} shows qualitative results for 3 example
queries. The first column describes the target image and its caption. The
next columns show the rank of the correct image and the top-4
ranked images for MIL\_STEM and CN\_MAX. Green words on the query are
stem-detectable and red words are CN-detectable but not stem-detectable.

Query 1 shows an example of images for which no detectors can be
used and thus the only piece of available information comes from
CN. Rank for MIL-STEM is, therefore, arbitrary and, as a result, the
correct image is under r@25. Query 2 shows an example where we
have both stem-detectable words and CN-detectable words (that are
not stem-detectable). In these cases, CN\_MAX is able to
detect ``\textit{bagel}'' using the ``\textit{doughnut}'' and
``\textit{bread}'' detectors (among others), improving the ranking of
the correct image. The last query is a case for which the CN-score is detrimental. 
For Query 3, the word ``\textit{resort}'' is highly related with ``\textit{hotel}''
in both CN and \esp, thus the "\textit{hotel}" detector became more
relevant than ``\textit{looking}'' and ``\textit{hills}''.

\begin{figure*}
\centering
\setlength{\tabcolsep}{0pt}
\begin{tabular}{| >{\centering}m{5cm}<{\centering} | m{12.12cm} | }
\hline
\input{./ex/header} \\\hline
\input{./ex/ex2} \\\hline
\input{./ex/ex3} \\\hline
\input{./ex/ex6} \\\hline
\end{tabular}
\caption{Qualitative examples for our baseline "MIL\_STEM" and our
  method "CN\_MAX" over COCO 5K. Green words are stemming-detectable,
  whereas red words are only CN-detectable. }
\label{fig:qualitative}
\end{figure*}

% Word detectors
Finally, we wanted to evaluate how good is the performance when
focusing only on those words that are CN-detectable but not
stemming-detectable. To that end, we design the following
experiment: we consider the set of words from the union of text tags that
are only CN-detectable, and we interpret those as one-word
queries. An image is ground truth in this case if any of its tags
contains $w$. 

Results in Table~\ref{tab:mscoco5k_words} are disaggregated for
word type (Nouns, Verbs, Adjectives). As a reference of the problem
difficulty, we add a random baseline. The results suggest that CN
yields more benefits for nouns, which may be easier to detect than
verbs and adjectives by CN\_MAX.

We observe that numbers are lower than in
Table~\ref{tab:mscoco5k_subset}. In part this is due to the fact that in
this experiment there is more than one correct image, therefore the
$k$ recall has higher chances of being lower than when there is only
one correct image. Furthermore, a qualitative look at the data suggests that
sometimes top-10 images are ``good'' even though ground truth images
were not ranked well. Figure~\ref{fig:qualitative_words} shows an
example of this phenomenon for the word \emph{tuxedo}.
%, which is not stemming-detectable.

%These results prompted us to experiment with a version of our CN score
%that tries to detect only undetectable nouns (but not verbs or
%adjectives). As we see below, better results are produced by this variant.

	\begin{table}\centering
          {\fs{9}
            \renewcommand{\arraystretch}{1.2} 
	\begin{tabular}{ | l | c c c c c |}
	\hline
	\textbf{Algorithm} & r@1 & r@5 & r@10 & median & mean  \\
	CN\_MAX &     &     &      & rank   & rank \\
	\hline
	Random & 0.02 & 0.1 & 0.2 & 2500.5 & 2500.50 \\
	\hline
	All & 0.4 & 1.8 & 3.3 & 962.0 & 1536.8 \\
	Noun & 0.5 & 2.1 & 3.7 & 755.0 & 1402.7 \\
	Verb & 0.2 & 1.1 & 1.9 & 1559.5 & 1896.2 \\
	Adjective & 0.1 & 0.7 & 1.9 & 1735.5 & 1985.2 \\
	\hline	
	\end{tabular}}
	\caption{Image Retrieval for new word detectors over COCO 5K. We include a random baseline, and results for CN\_MAX divided in 4 categories: Retrieving nouns, verbs, adjectives and all of them. The results show that is easier, for CN\_MAX, to detect nouns than verbs or adjectives. }
	\label{tab:mscoco5k_words}
	\end{table}

\begin{figure}
\centering
\setlength{\tabcolsep}{0pt}
\begin{tabular}{| m{7.7cm} | }
\hline
\textbf{Target word}: Tuxedo \\\hline
\input{./ex_w/ex1} \\\hline
\end{tabular}
\caption{ Qualitative examples for \textit{tuxedo} retrieval. First image row contains our ground truth, the 4 examples where \textit{tuxedo} was used to describe the image. The second row of images are the first 4 retrieved images from CN\_MAX. }
\label{fig:qualitative_words}
\end{figure}

% ESPGAMES WAS KEY
\Paragraph{Visual knowledge from \esp is key} 
We experiment with three
alternative ways to compute CN-detectable scores which do not
yield good results. First, we use CN considering
$\Phat(w | w',I) = 1$ and $\Phat(w |\neg w',I) = 0$ for
Equation~\ref{eq:phatww}. We also try estimating
$\Phat(w | w',I)$ using CN weights. Finally, we use the
similarity measure of Word2Vector \cite{Mikolov:EtAl:2013} as an
alternative to \esp. All those variants perform worse
than ESP\_MEAN\_G.

% Other types of relationships
\Paragraph{Considering the relation types} 
We explore using different subsets of relation types, but the performance always decreases. To our surprise, even removing the "Antonym" relation decreases the overall performance. We study this phenomenon and discover that a high number of antonym relationships are visually relevant (e.g. "cat \rel{Antonym} dog"). As CN is built by people, we believe that even non-visual relation types, such as "Antonym", are biased towards concepts that are somehow related in our mind. This might partially explain why the best performance is obtained by letting \esp to choose visually relevant relation instances without considering their type. Nonetheless, we believe that relation types are a valuable information that we have not discovered how to properly use.

% Comparison to previous works
\subsection{Comparison to  Other Approaches}
We compare with NeuralTalk\footnote{%We used %the source code from
  \url{https://github.com/karpathy/neuraltalk}.} 
\cite{Bengio:Caption:2015}, 
BRNN \cite{karphati:Caption:2015}, and GMM+HGLMM (the best
algorithm in \cite{KleinLSW15}) over COCO~5K.  To reduce the impact of
noisy relations in ConceptNet in CN\_MAX, we only consider
relationships with CN confidence weight $\geq 1$ (this threshold is
defined by carrying out a sensitivity analysis). As we can see on
table~\ref{tab:mscoco5k}, MIL outperforms previous approaches to image
retrieval. Moreover, adding CN to detect new words improves
the performance in almost all metrics. 

% Database
% Some decision + sensibility

\begin{table} \centering
          {\fs{9}
            \renewcommand{\arraystretch}{1.2} 
   \setlength{\tabcolsep}{4pt}
  \begin{tabular}{|l|ccccc|}
    \hline
    \textbf{Database} & r@1 & r@5 & r@10 & median & mean \\
    COCO 5K &     &     &      & rank   & rank \\
    \hline	
    \textbf{Other approaches} & & & & & \\
    % Mean Vec  & 10.2 & 27.2 & 38.3 & 18 & 64.8 \\
    % GMM       & 10.2 & 27.7 & 39.4 & 17 & 51.1 \\
    % LMM       & 10.5 & 28.0 & 40.0 & 17 & 51.3 \\
    % HGLMM     & 10.4 & 27.7 & 39.7 & 17 & 50.7 \\
    %NeuralTalk  [Karp. et al , 2015; Vin. et al., 2015]\cite{karphati:Caption:2015,Bengio:Caption:2015} & 6.9 & 22.1 & 33.6 & 22 & 72.2 \\
    NeuralTalk & 6.9 & 22.1 & 33.6 & 22 & 72.2 \\
    GMM+HGLMM & 10.8 & 28.3 & 40.1 & 17 & 49.3 \\
    BRNN      & 10.7 & 29.6 & 42.2 & 14 & NA \\
    \hline
    \textbf{Our baselines}   & & & & & \\
    MIL       & 15.7 & 37.8 & 50.5 & \textbf{10} & 53.6 \\	
    MIL\_STEM & 15.9 & 38.3 & 51.0 & \textbf{10} & 49.9 \\
    \hline
    \textbf{Our method}   & & & & & \\
    CN\_MAX & \textbf{16.2} & \textbf{39.1} & \textbf{51.9} & \textbf{10} & \textbf{44.4} \\
    \hline	
  \end{tabular}}
  \caption{Image retrieval results over COCO 5K. References: NeuralTalk \protect\cite{Bengio:Caption:2015},  BRNN \protect\cite{karphati:Caption:2015}, and GMM+HGLMM \protect\cite{KleinLSW15}.}%Note our baselines outperform previous works and our approach, CN\_MAX, improved our baselines performance. 
  \label{tab:mscoco5k}
\end{table}

\subsection{From COCO 5K to  COCO 22K}
We test our method on 22K images of the MS~COCO database. With more
images, the difficulty of the task increases. 
%In preliminary experiments we did not find CN\_MAX to improve performance; however,
Motivated by the fact that CN seems to be best at noun detection (\cf,
Table~\ref{tab:mscoco5k_words}), we design   a version of CN\_MAX,
called CN\_MAX (NN), whose CN-score only focuses on undetectable nouns.

Table~\ref{tab:mscoco40k} shows the results for MIL\_STEM and CN\_MAX
(NN). We show results over COCO 5K and 22K. Interestingly, the
improvement of CN\_MAX over MIL\_STEM \emph{increases} when we add
more images. Notably, we improve upon the median score, which is a
good measure of a significant improvement.

%Intuitively, with more images there are more target queries with undetectable words. Thus every reduction added by ConceptNet 

% Adding more images
	
	\begin{table}\centering
          {\fs{9}
            \renewcommand{\arraystretch}{1.2} 
	\begin{tabular}{ | l | c c c c c |}
	\hline
	\textbf{Databases} & r@1 & r@5 & r@10 & median & mean  \\
	From 5K to 22K &     &     &      & rank   & rank \\
	\hline
	\textbf{COCO 5K} &     &     &      &    &  \\
	MIL\_STEM & 15.9 & 38.3 & 51.0 & 10 & 49.9 \\
	CN\_MAX (NN) & 16.3 & 39.2 & 51.9 & 10 & 44.5 \\
	Improvement (\%) & \textbf{2.5} & 2.4 & 1.8 & 0 & 10.8 \\		
	\hline
	\textbf{COCO 22K} &     &     &      &    &  \\
	MIL\_STEM &	7.0 & 18.7 & 26.6 & 43 & 224.6 \\
	CN\_MAX  & 7.1 & 19.1 & 27.1 & 42 & 198.8 \\
	CN\_MAX (NN) & 7.1 & 19.2 & 27.2 & 41 & 199.7 \\
	Improvement (\%) & 1.4 & \textbf{2.7} & \textbf{2.3} & \textbf{5} & \textbf{46.7}  \\
	%\hline
	%\textbf{MSCoco 40K} &     &     &      &    &  \\
	%MIL\_STEM &	 &  &  &  &  \\
	%CN\_MAX  &  &  &  &  &  \\
	%CN\_MAX  (NN) &  &  &  &  &  \\
	%Improvement (\%) & \textbf{??} & \textbf{??} & \textbf{??} & \textbf{??} & \textbf{??} \\
	\hline
	\end{tabular}}
      \caption{Image retrieval results for COCO 5K and 22K. In this
        table we compare our best baseline against a version of
        CN\_MAX which only detects new noun words. Performance improvement increases when more images are considered. }	
	\label{tab:mscoco40k}	
	\end{table}

% \subsection{The Power of ConceptNet}

% \begin{itemize}
% \item Why did conceptnet improve the results?
% \begin{itemize}
% \item ConceptNet acts as a regularizator for ESPGAMES probabilities, preventing it from overfitting.
% \end{itemize}
% \item Why did ConceptNet improve the results more in 22K?
% \begin{itemize}
% \item More hard examples?
% \end{itemize}
% \end{itemize}

% %. We think that this happen because ConceptNet acts as a regularizator for equation \textbf{(?)}, preventing it to overfit. We believe that common sense ontologies are a useful regularizator for machine learning techniques, and this experiment support our belief.

%% file: ex/header.tex
Sentence and target image & 
\setlength{\tabcolsep}{6pt}
\begin{tabular}{ >{\centering}b{2cm}<{\centering} >{\centering}m{2cm}<{\centering} >{\centering}m{2cm}<{\centering} >{\centering}m{2cm}<{\centering} >{\centering}m{2cm}<{\centering} }
Algorithm & Pos 1 & Pos 2 & Pos 3 & Pos 4 %\\ \hline
\end{tabular}

%% file: ex/ex2.tex
\includegraphics[width=3cm,height=2cm,keepaspectratio]{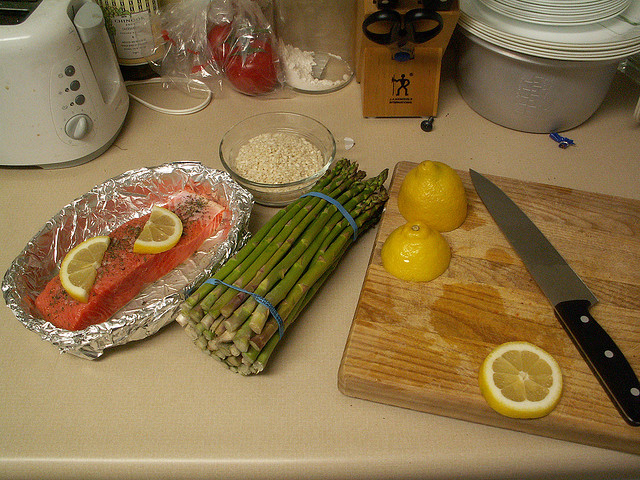}

1) The \textcolor{cn}{preparation} of \textcolor{cn}{salmon}, \textcolor{cn}{asparagus} and \textcolor{cn}{lemons}.
& 
\setlength{\tabcolsep}{6pt}
\begin{tabular}{ >{\centering}m{2cm}<{\centering} m{2cm} m{2cm} m{2cm} m{2cm}  }
MIL\_STEM Pos: - &
\includegraphics[width=2cm,height=1.5cm,keepaspectratio]{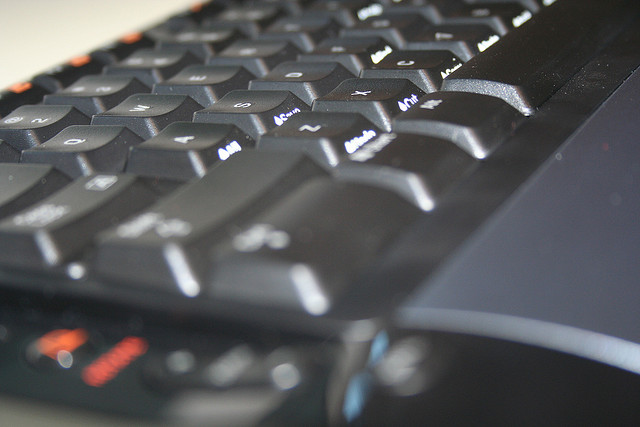} & 
\includegraphics[width=2cm,height=1.5cm,keepaspectratio]{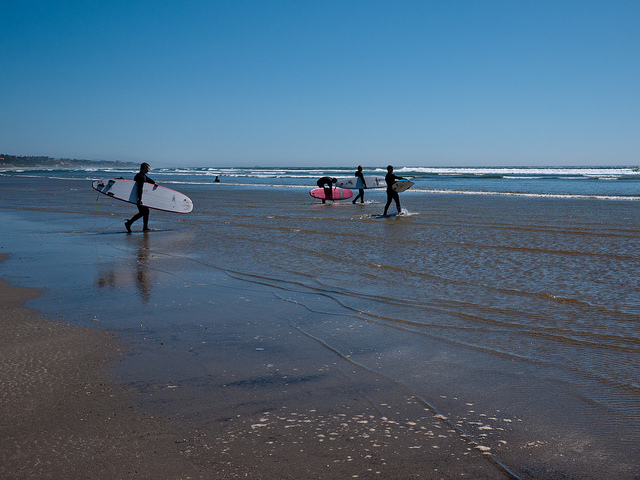} & 
\includegraphics[width=2cm,height=1.5cm,keepaspectratio]{./img/1_3b.jpg} &
\includegraphics[width=2cm,height=1.5cm,keepaspectratio]{./img/1_4b.jpg} \\
\hline
CN\_MAX Pos: \textbf{23} & 
\includegraphics[width=2cm,height=1.5cm,keepaspectratio]{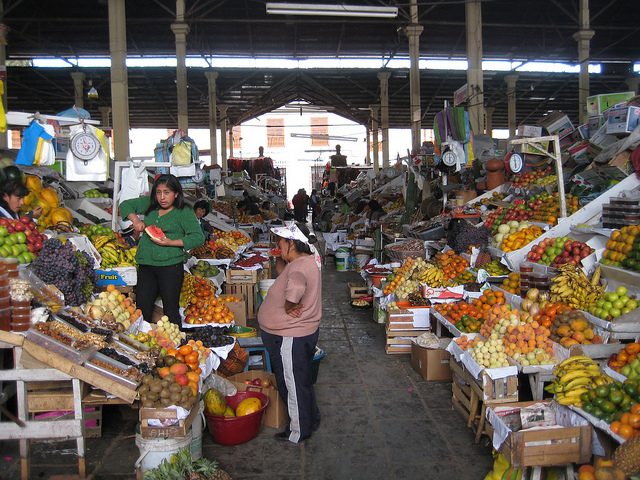} &
\includegraphics[width=2cm,height=1.5cm,keepaspectratio]{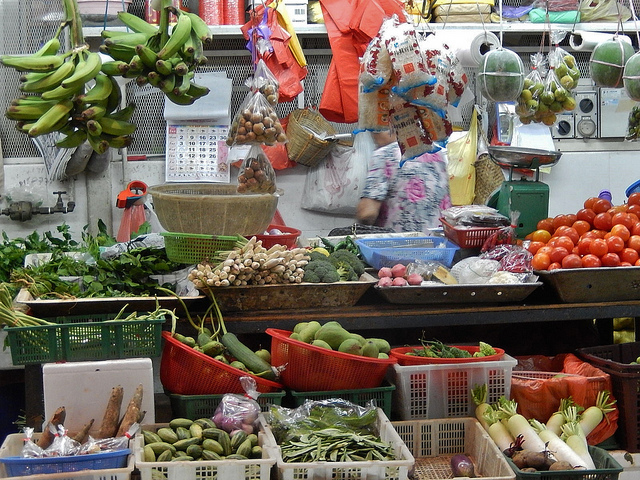} &
\includegraphics[width=2cm,height=1.5cm,keepaspectratio]{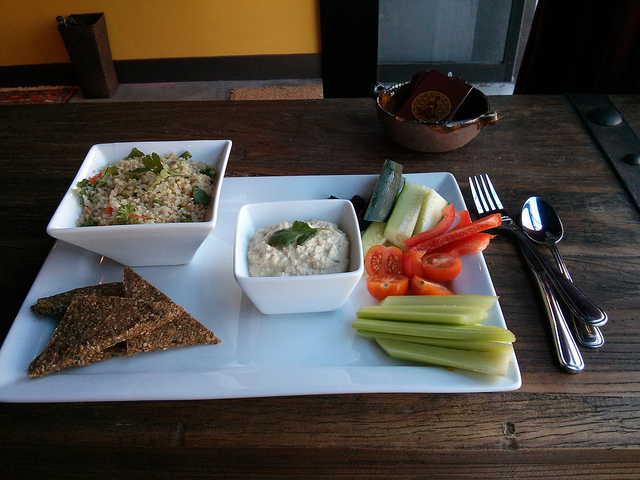} &
\includegraphics[width=2cm,height=1.5cm,keepaspectratio]{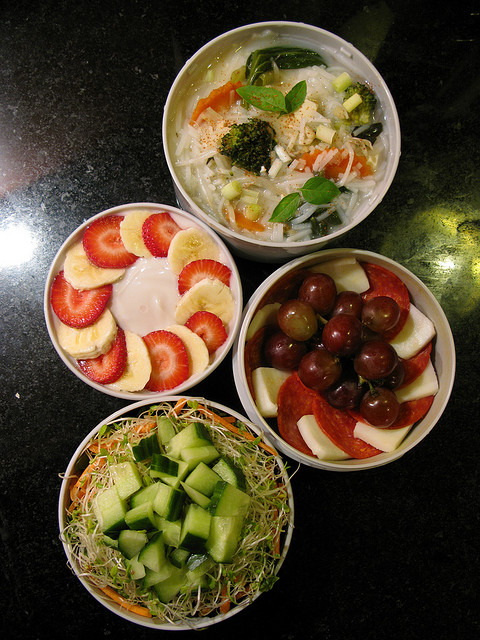} \\
\end{tabular}

%% file: ex/ex3.tex
\includegraphics[width=3cm,height=2cm,keepaspectratio]{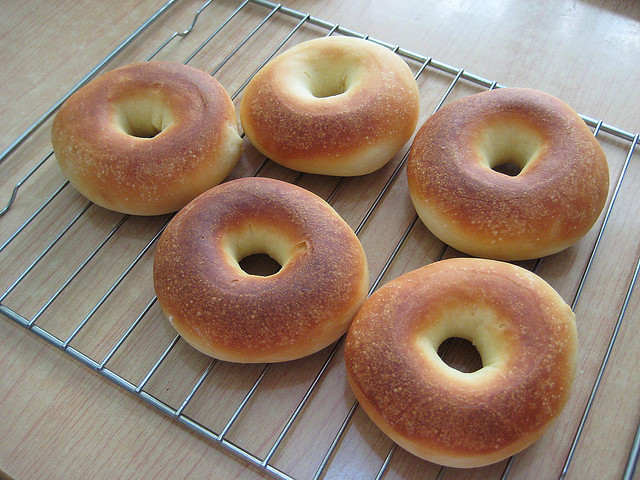}

2) Those \textcolor{cn}{bagels} \textcolor{mil}{are} \textcolor{mil}{plain} with \textcolor{cn}{nothing} on them.
& 
\setlength{\tabcolsep}{6pt}
\begin{tabular}{ >{\centering}m{2cm}<{\centering} m{2cm} m{2cm} m{2cm} m{2cm} m{2cm} }
MIL\_STEM Pos: 360 &
\includegraphics[width=2cm,height=1.5cm,keepaspectratio]{./img/3_1b.jpg} & 
\includegraphics[width=2cm,height=1.5cm,keepaspectratio]{./img/3_2b.jpg} & 
\includegraphics[width=2cm,height=1.5cm,keepaspectratio]{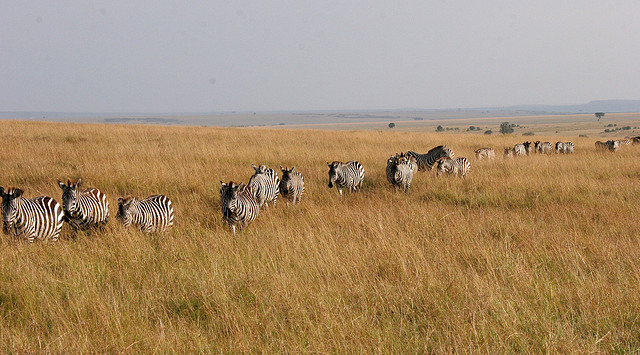} &
\includegraphics[width=2cm,height=1.5cm,keepaspectratio]{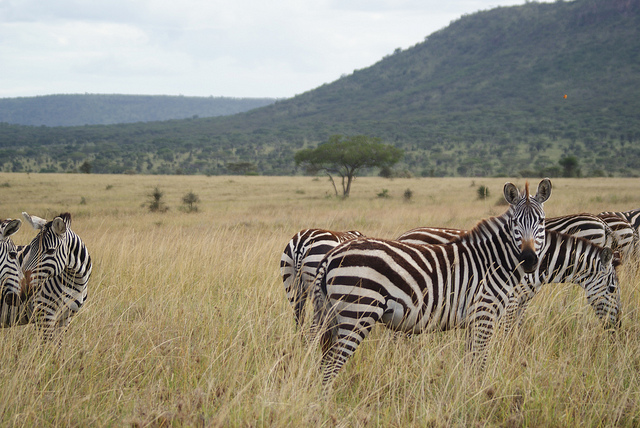} \\
\hline
CN\_MAX Pos: \textbf{2} & 
\includegraphics[width=2cm,height=1.5cm,keepaspectratio]{./img/3_1.jpg} &
\includegraphics[width=2cm,height=1.5cm,keepaspectratio]{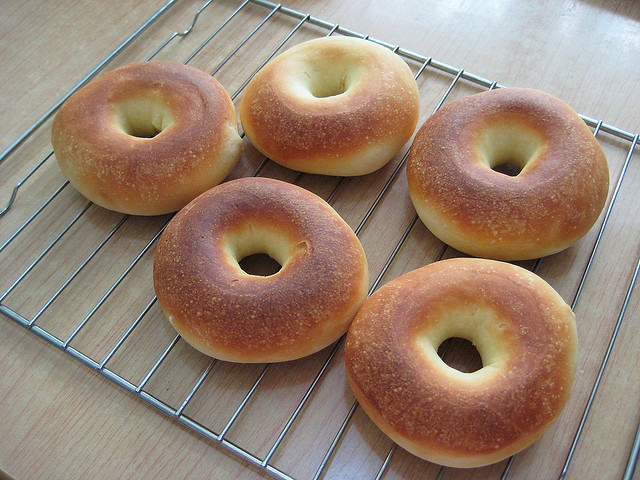} &
\includegraphics[width=2cm,height=1.5cm,keepaspectratio]{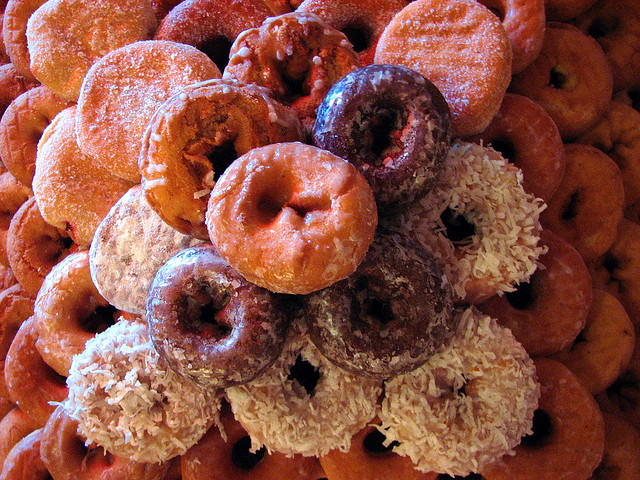} &
\includegraphics[width=2cm,height=1.5cm,keepaspectratio]{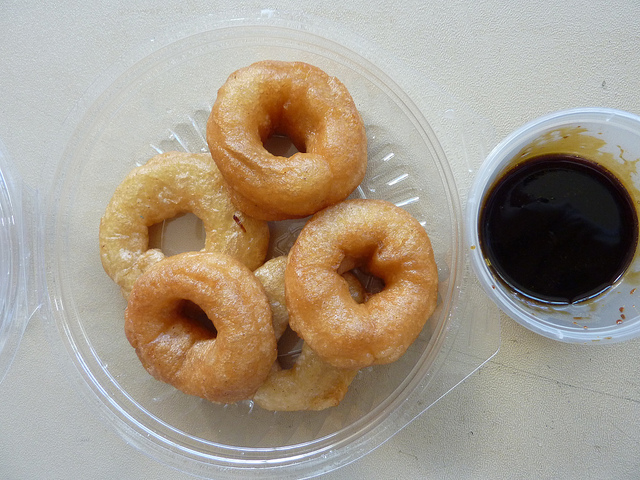} \\
\end{tabular}

%% file: ex/ex6.tex
\includegraphics[width=3cm,height=2cm,keepaspectratio]{./img/6.jpg}

3) A spooky \textcolor{mil}{looking} \textcolor{mil}{hotel}\\ \textcolor{cn}{resort} in the \textcolor{mil}{hills}.& 
\setlength{\tabcolsep}{6pt}
\begin{tabular}{ >{\centering}m{2cm}<{\centering} m{2cm} m{2cm} m{2cm} m{2cm} m{2cm} }
MIL\_STEM Pos: \textbf{349} &
\includegraphics[width=2cm,height=1.5cm,keepaspectratio]{./img/6_1b.jpg} & 
\includegraphics[width=2cm,height=1.5cm,keepaspectratio]{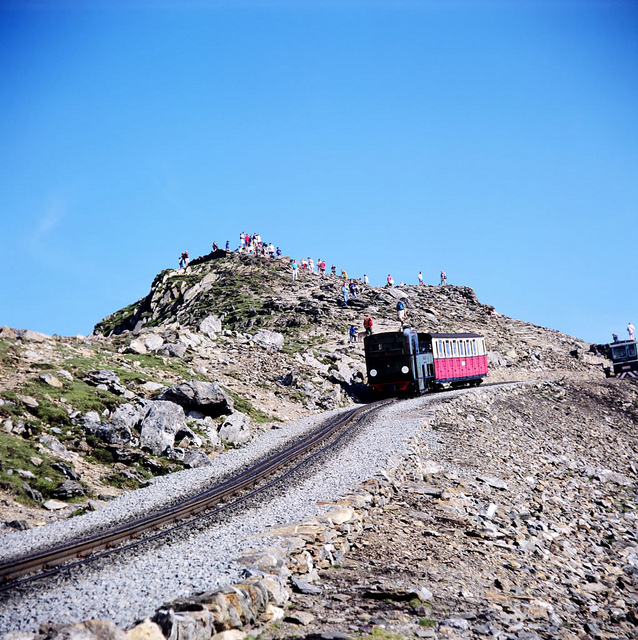} & 
\includegraphics[width=2cm,height=1.5cm,keepaspectratio]{./img/6_3b.jpg} &
\includegraphics[width=2cm,height=1.5cm,keepaspectratio]{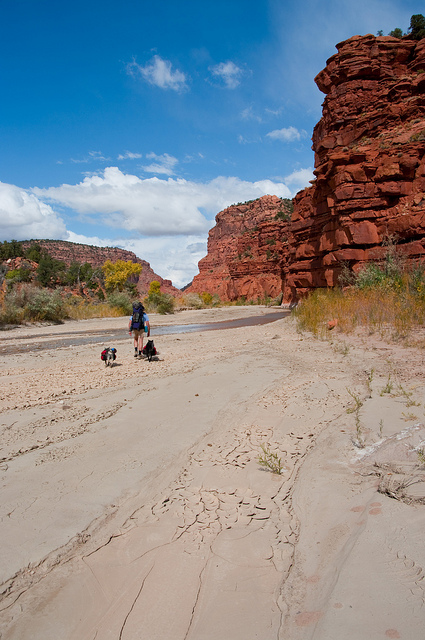} \\
\hline
CN\_MAX Pos: 597 & 
\includegraphics[width=2cm,height=1.5cm,keepaspectratio]{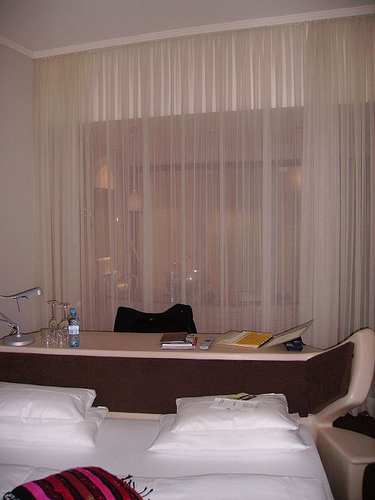} &
\includegraphics[width=2cm,height=1.5cm,keepaspectratio]{./img/6_2.jpg} &
\includegraphics[width=2cm,height=1.5cm,keepaspectratio]{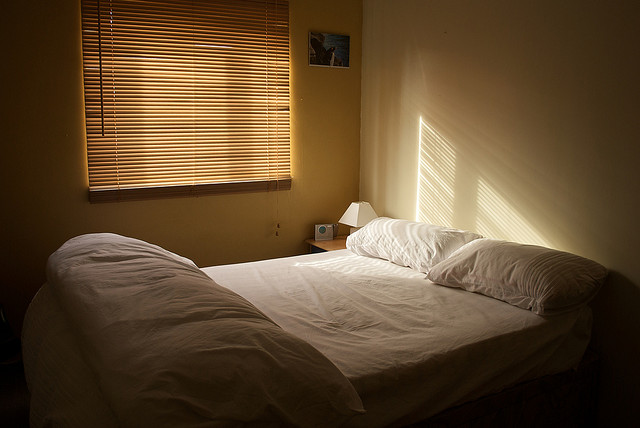} &
\includegraphics[width=2cm,height=1.5cm,keepaspectratio]{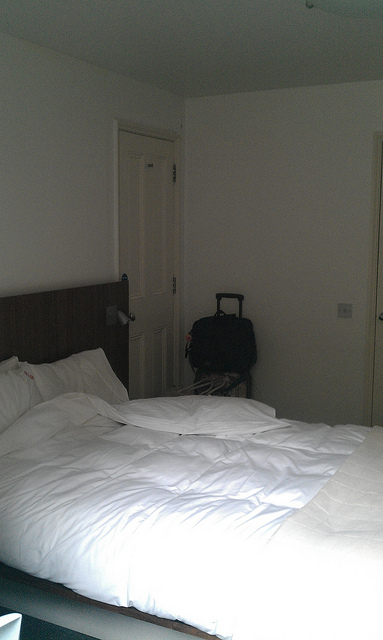} \\
\end{tabular}

%% file: ex_w/ex1.tex
Ground truth images positions: 1, 130, 192 and 275 \\ \hline
\setlength{\tabcolsep}{6pt}
\begin{tabular}{ m{1.5cm} m{1.5cm} m{1.5cm} m{1.5cm} m{1.5cm} }
\includegraphics[width=1.5cm,height=1.5cm,keepaspectratio]{./img_words/1_1b.jpg} & 
\includegraphics[width=1.5cm,height=1.5cm,keepaspectratio]{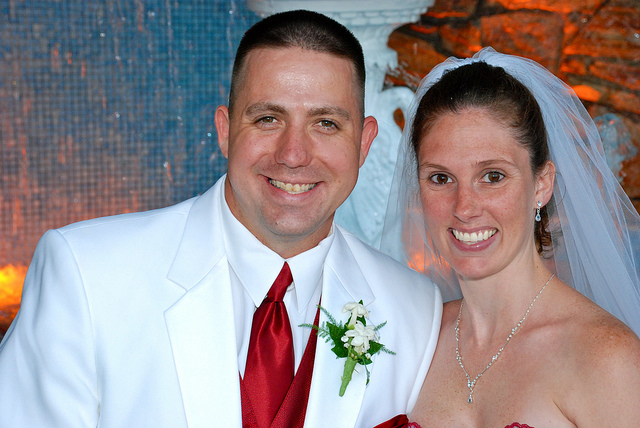} & 
\includegraphics[width=1.5cm,height=1.5cm,keepaspectratio]{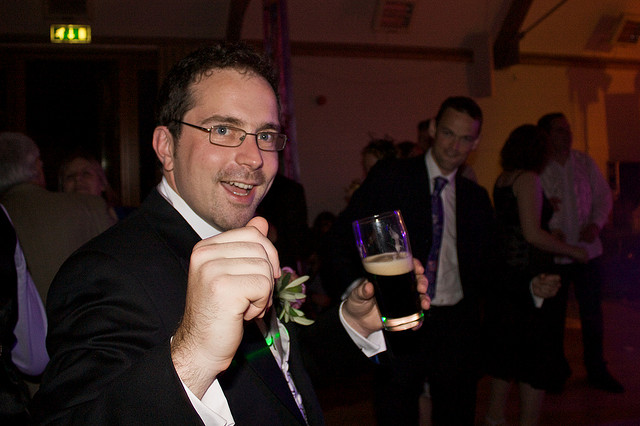} &
\includegraphics[width=1.5cm,height=1.5cm,keepaspectratio]{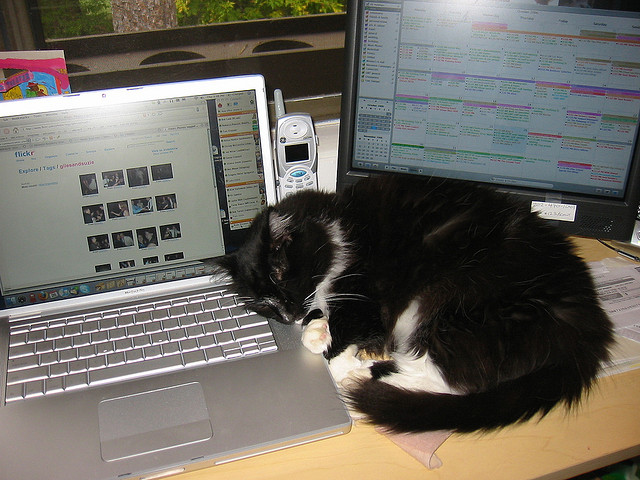} \\
\end{tabular} \\ \hline
Retrieved images position 1, 2, 3 and 4 \\ \hline
\setlength{\tabcolsep}{6pt}
\begin{tabular}{ m{1.5cm} m{1.5cm} m{1.5cm} m{1.5cm} m{1.5cm} }
\includegraphics[width=1.5cm,height=1.5cm,keepaspectratio]{./img_words/1_1.jpg} &
\includegraphics[width=1.5cm,height=1.5cm,keepaspectratio]{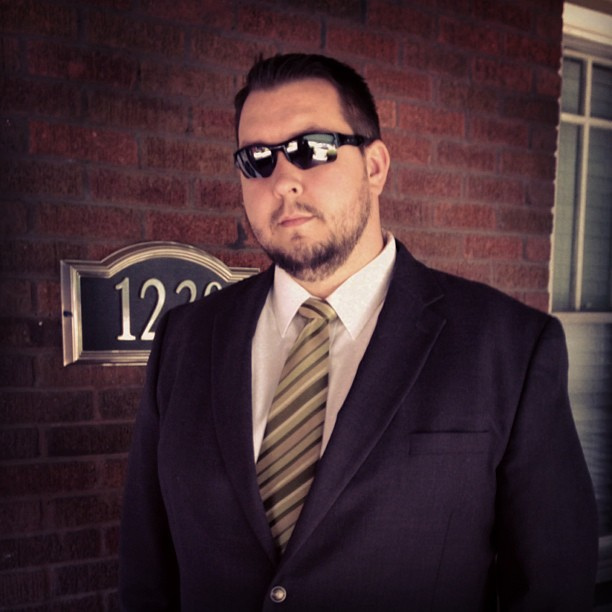} &
\includegraphics[width=1.5cm,height=1.5cm,keepaspectratio]{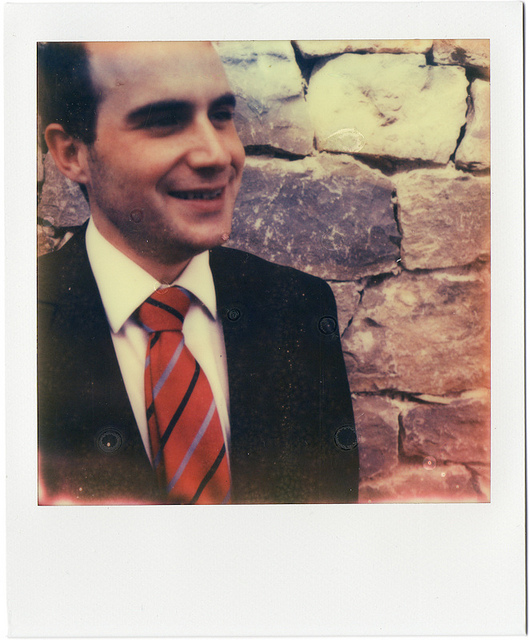} &
\includegraphics[width=1.5cm,height=1.5cm,keepaspectratio]{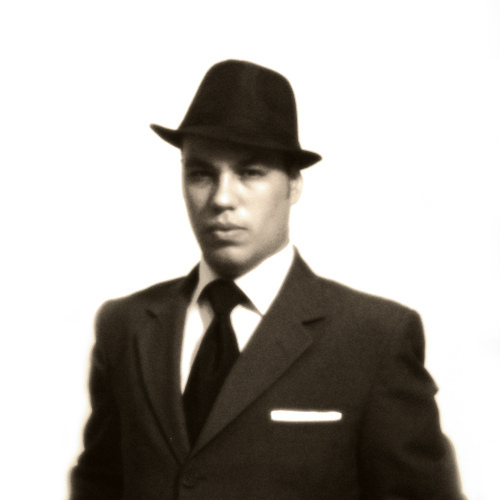} \\
\end{tabular}

%% file: sections/conclusions.tex
\section{Conclusions and Perspectives}\label{Sec:Conclusions}
This paper presented an approach to enhancing a learning-based
technique for sentence-based image retrieval with general-purpose
knowledge provided by ConceptNet, a large commonsense ontology. Our
experimental data, restricted to the task of image retrieval, shows
improvements across different metrics and experimental 
settings. This suggests a promising research area where the benefits of 
integrating the areas of knowledge representation and computer vision 
should continue to be explored.

%This paper presented an approach to enhancing a learning-based
%technique for sentence-based image retrieval with general-purpose
%knowledge provided by ConceptNet, a large commonsense ontology. Our
%experimental data, restricted to the task of image retrieval, shows
%improvements leading to state-of-the-art performance. This suggests
%a promising research area where the benefits of integrating
%the areas of knowledge representation and computer vision should
%continue to be explored.

An important conclusion of this work is that
integration of a general-purpose ontology with a vision approach is
not straightforward. This is illustrated by the experimental data that
showed that information in the ontology alone \emph{did not} improve
performance, while the \emph{combination} of an ontology and
crowd-sourced visual knowledge (from \esp) did. This suggests that future
works in the intersection of knowledge representation and vision may require 
special attention to relevance and knowledge base filtering.

%There are many perspectives for future work. For example,
%an issue we did not address is one related to the polysemy and
%synonymy issues. For example, in ConceptNet the node \textit{fly} has
%relations for the \textit{verb fly} and for the \textit{insect fly};
%meanwhile, \textit{computer} has relations to concepts that
%\textit{pc} does not. The stemming process---that we needed to use in
%order to apply CN---may sometimes introduce ambiguity; for example, by
%transforming an undetectable word such as \textit{bowling} into
%\textit{bowl}. Our approach also can be further optimized for the task
%of image retrieval. 
%For example, currently to detect \textit{tuxedo},
%we look for images with high scores for \textit{jacket},
%\textit{black}, etc., but we do not constrain these detectors to fire
%at the same location.
%Second, sentence understanding is key; sentences like
%``\textit{a road without cars}'' produce really bad image retrieval
%responses. Finally, our score is a simple multiplication of the
%detectors outputs but, empirically, we have seen that query sentences
%usually have words that are more important to detect than others.